\documentclass[conference]{IEEEtran}

\usepackage[pdftex]{graphicx}
\usepackage[ruled,vlined,linesnumbered]{algorithm2e}
\usepackage{array}
\usepackage{hyperref}

\AtBeginDocument{%
  \providecommand\BibTeX{{%
    \normalfont B\kern-0.5em{\scshape i\kern-0.25em b}\kern-0.8em\TeX}}}

\begin{document}

\newenvironment{conditions}
  {\par\vspace{\abovedisplayskip}\noindent\begin{tabular}{>{$}l<{$} @{${}={}$} l}}
  {\end{tabular}\par\vspace{\belowdisplayskip}}

\title{Deep Evolution for Facial Emotion Recognition}

\author{\IEEEauthorblockN{Emmanuel Dufourq}
\IEEEauthorblockA{African Institute for Mathematical Sciences \\ \ Stellenbosch University\\
Email: edufourq@gmail.com}
\and
\IEEEauthorblockN{Bruce A. Bassett}
\IEEEauthorblockA{African Institute for Mathematical Sciences\\ South African Astronomical Observatory \\ South African Radio Astronomy Observatory \\ Department of Maths and Applied Maths, University of Cape Town \\ 
Email: bruce.a.bassett@gmail.com}
}
\maketitle

\begin{abstract}
Deep facial expression recognition faces two challenges that both stem from the large number of trainable parameters: long training times and a lack of interpretability. We propose a novel method based on evolutionary algorithms, that deals with both challenges by massively reducing the number of trainable parameters, whilst simultaneously retaining classification performance, and in some cases achieving superior performance. We are robustly able to reduce the number of parameters on average by  95\% (e.g. from 2M to 100k parameters) with no loss in classification accuracy. The algorithm learns to choose small patches from the image, relative to the nose, which carry the most important information about emotion, and which coincide with typical human choices of important features. Our work implements a novel form 
attention and shows that evolutionary algorithms are a valuable addition to machine learning in the deep learning era, both for reducing the number of parameters for facial expression recognition and for providing interpretable features that can help reduce bias. 
\end{abstract}

\begin{IEEEkeywords}
convolutional neural networks, evolutionary algorithms, facial expression recognition, image classification, neural network compression
\end{IEEEkeywords}

\section{Introduction}

Humans convey their emotions across in various forms, one of which is expressed through changes in a person's face as a consequence of their emotional state \cite{Tian:2011:fFER}. There are countless numbers of interactions between humans each day given that humans are a social species. The human face provides a lot of information in these interactions \cite{Jack:2015:TheHuman}. Facial Expression Recognition (FER) is the ability to recognise the expressions that are being conveyed through the changes in the face.

There has been a lot of work in this field of research given the progress in computer vision research, see \cite{Calenau:2013:FERABrief} for a review of works from 2003 to 2012. Amongst the vast available set of machine learning algorithms, there has been a tremendous amount of recent work using deep learning \cite{LeCun:2015:DeepLearning}. Convolutional neural networks (CNN) \cite{LeCun:1990:HandwrittenDigit} are widely used in image classification tasks. Researchers focusing on FER have since been using CNNs as they achieve higher classification performance in comparison to earlier methods. 

There are existing literature reviews on the topic of machine learning and FER. Corneanu \textit{et al.} \cite{Corneanu:2016:SurveyonRGB} provides a useful taxonomy of FER and computer vision research areas such as face localisation, feature extraction, classification and multimodal fusion -- these are discussed in the context of RGB, thermal and 3D images. In terms of the classification areas not a lot of CNN studies were analysed. Pramerdorfer and Martin \cite{pramerdorfer:2016:FERUsing} compared 6 CNN studies of depths 5 to 11 against VGG, Inception, ResNet and an ensemble on a single dataset. Pantic and Rothkrantz \cite{Pantic:2000:AutomaticAnalysis} discussed some neural network studies but did not focus on CNNs, similarly, Sariyanidi \textit{et al.} and Wu \textit{et al.} \cite{Sariyanidi:2015:AutomaticAnalysis, Wu:2012:SurveyOf} did not describe CNN related works. Martinez \textit{et al.}  \cite{Martinez:2016:AdvancesChallenges} reviewed a number of studies and described challenges and opportunities in FER research such as medical, marketing and audience monitoring as well as human computer interaction studies. They did not review studies that implemented CNNs. Ghayoumi \cite{ghayoumi:2017:AQuickReview} discussed a few CNN studies. Zhang \cite{Zhang:2018:FERBased} discussed a few deep belief networks and CNNs. Ko \cite{Ko:2018:ABriefReview} describe a number of long short term memory CNN approaches which have been applied to FER. Latha and Priya \cite{Latha:2016:AReviewOnDeep} discuss 17 CNN studies which have been applied to FER. 

It is clear from the existing work in the literature that CNNs achieve state-of-the-art performance on FER tasks as opposed to other machine learning methods. Training CNNs often result in a large number of trainable parameters. This in turn implies that long training times are to be expected on limited hardware. In this study, we explore a novel idea which attempts to optimise the predictive performance of CNNs for FER and simultaneously, reduce the number of neural network trainable parameters without compromising on the classification accuracy. We ask the following, can we achieve similar predictive performance when training a model on small image patches (extracted from an image) as opposed to the entire image of the face? Are certain facial features more discriminative for FER and can an evolutionary algorithm \cite{Miller:1996:GAS:1326713.1326715} discover these areas? Can an evolutionary algorithm optimise the size of small image patches to reduce the input image size and consequently the number of trainable CNN parameters?

\section{Proposed Approach}\label{sec:evofer-proposed}

Based on the literature surveyed, there have been no prior attempts at using an evolutionary algorithm to extract patches from images with the objective of reducing the number of trainable CNN parameters and to retain (or improve) classification performance. We propose \textit{Evolutionary Facial Expression Recognition} (EvoFER) in this section. \textit{EvoFER} extracts a number of patches from the original image and stacks the patches to form a new image. 

\begin{figure}
  \centering
          \includegraphics[width=0.48\textwidth]{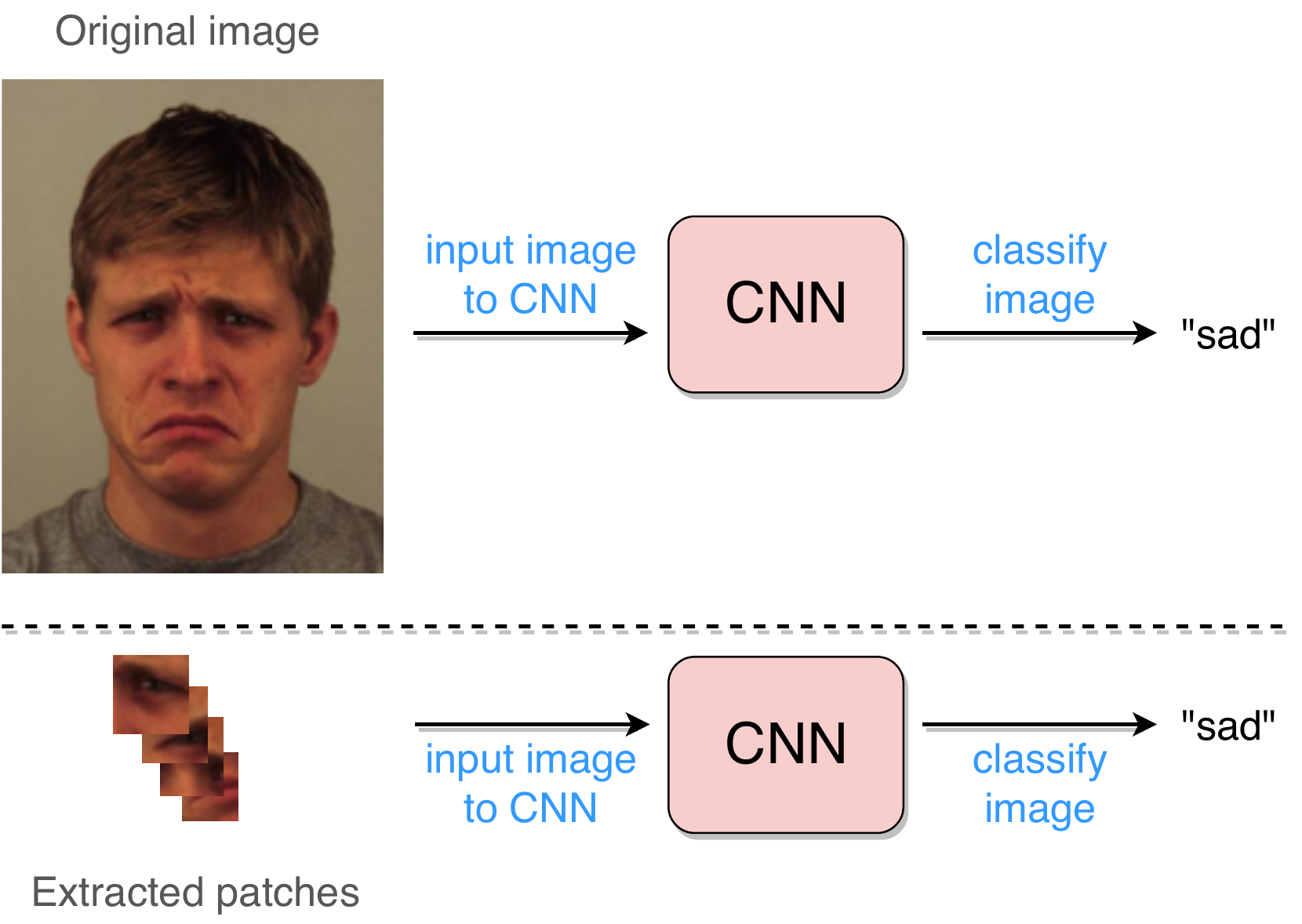}
  \caption[Illustrating the primary idea behind EvoFER.]{Illustrating the primary idea behind \textit{EvoFER}. Instead of inputting an entire face image into a CNN to get the prediction, we propose to input patches to achieve similar predictive performance. The extracted patches are stacked together to form a new image. Expression image extracted from \cite{KDEF:1998:KDEF}.} \label{fig:evofer-patches-to-cnn}
\end{figure}

Figure \ref{fig:evofer-patches-to-cnn} illustrates an example whereby an image is input into a CNN for classification, and below, an example of four stacked patches (extracted from \textit{EvoFER}) being inputted into the same CNN. After converting to greyscale, the resolution of the original image is \textit{281$\times$381$\times$1} whereas the new stacked image is  \textit{50$\times$50$\times$4}. Thus, the number of trainable CNN parameters using the original image is greater than the number of parameters when using the new stacked image made up of several patches. The patches are stacked in a similar manner to a colour image having 3 colour channels (RGB).

The objective function for \textit{EvoFER} is multi-objective. Firstly, \textit{EvoFER} attempts to reduce the number of trainable parameters as opposed to the number of parameters which would be used when training on the full image. The second objective is to achieve the highest possible classification accuracy using the patches as opposed to using the full image. The following subsections describe \textit{EvoFER}. 

\subsection{Chromosome}

\begin{figure}
  \centering
          \includegraphics[width=0.48\textwidth]{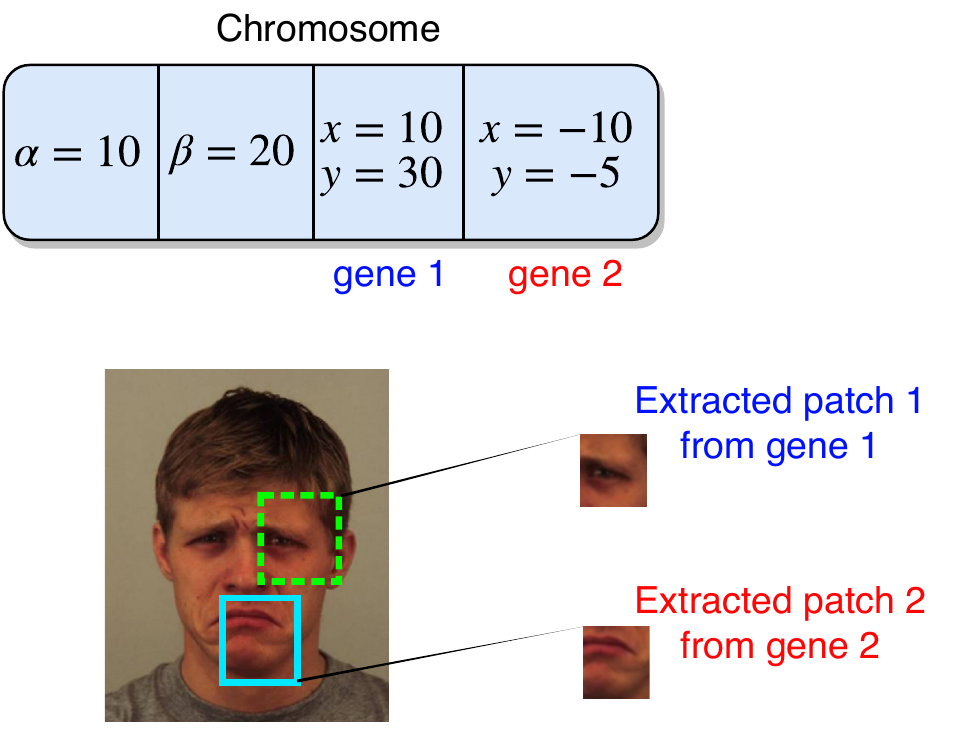}
  \caption[Illustrating a EvoFER chromosome which extracts two patches.]{Illustrating a \textit{EvoFER} chromosome which extracts two patches. The width of the patches to extract is 10 and the height is 20. The figure illustrates how the chromosome is applied to an image to extract a patch. Expression image extracted from \cite{KDEF:1998:KDEF}.} \label{fig:evofer-chromosome}
\end{figure}

\textit{EvoFER} chromosomes contain 2 fixed genes and then allows for a number of pair of genes to be added. The first two genes denote values $\alpha$ and $\beta$ which represent the width and height of the patches to be extracted. The remainder of the chromosome denotes $(x, y)$ pairs. Each $x$ and $y$ pair are the coordinates of the top left point of the corresponding patch to be extracted from the original image. The coordinates $x$ and $y$ are relative to the location of the nose. The nose is used as a reference point and was obtained using OpenCV and DLIB \footnote{http://dlib.net/imaging.html}. The patch is extracted by obtaining the coordinate $(x, y)$ and using $\alpha$ and $\beta$ to obtain the entire patch. A chromosome thus encodes the location and size of the patches to be extracted from the original images. A chromosome must have at least one $(x, y)$ pair. User-defined parameters specify the maximum number of $(x, y)$ pairs allowed in each chromosome. The patches are stacked on top of each other vertically which in turn creates a new image. The order of the patches is not important. Figure \ref{fig:evofer-chromosome} illustrates an example of a chromosome with two patches. The width of the patches to be extracted is 10 and the height is 20. The green dotted line is the first patch and the solid blue line is the second patch. The location of the top left corner of each patch to extract relative to the coordinates of the nose are (10, 30) and (-10, -5). The figure illustrates the unstacked exacted patches.

\subsection{Initial Population Generation}

We use the standard initial population generation algorithm. We propose algorithm \ref{algo:create-evo-chromosome} to generate the \textit{EvoFER} chromosomes. This algorithm is executed a number of times based on the number of chromosomes to create. A random value for $\alpha$ and $\beta$ is assigned to each chromosome based on a corresponding bound which is pre-defined by the experimenter. The values of $\alpha$ and $\beta$ are not modified during the evolutionary process. 

\begin{algorithm}
\SetKwData{minAlpha}{min\_alpha}
\SetKwData{maxAlpha}{max\_alpha}
\SetKwData{minBeta}{min\_beta}
\SetKwData{maxBeta}{max\_beta}
\SetKwData{minPatches}{min\_patches}
\SetKwData{maxPatches}{max\_patches}
\SetKwData{patches}{patches}
\SetKwData{alpha}{alpha}
\SetKwData{beta}{beta}
\SetKwData{imageW}{image\_width}
\SetKwData{imageH}{image\_height}

\SetKwInOut{Input}{input}
\SetKwProg{Fn}{Function}{}{}

\Input{\minAlpha : minimum size for alpha}
\Input{\minBeta : minimum size for beta}
\Input{\maxAlpha : maximum size for alpha}
\Input{\maxBeta : maximum size for beta}
\Input{\minPatches : minimum number of patches allowed}
\Input{\maxPatches : maximum number of patches allowed}

	\Begin{
	
	Initialise an empty chromosome.

	\alpha $\leftarrow$ random[\minAlpha, \maxAlpha]

	\beta $\leftarrow$ random[\minBeta, \maxBeta]

	\patches $\leftarrow$ random[\minPatches, \maxPatches]

	\For{$i \gets 0$ \KwTo $\patches$}{

		X $\leftarrow$ $GenerateX(\alpha)$

		Y $\leftarrow$ $GenerateY(\beta)$

		Append $(X,Y)$ to chromosome

	}
	
\Return{chromosome.}

\Fn{GenerateX (\alpha)}{
\Return random[$-(\imageW / 2 - alpha)$, $\imageW / 2 - alpha$]
}

\Fn{GenerateY (\beta)}{
\Return random[$-(\imageH / 2 - beta)$, $\imageH / 2 - beta$]
}

}
\caption{Creating an EvoFER chromosome.} 
\label{algo:create-evo-chromosome} 
\end{algorithm}

Upon the creation of a chromosome, the number of patches is randomly assigned based on a user specified bound. For each patch to be created, a random value for $x$ and $y$ is created. These values are randomly generated based on the image width and height as is illustrated from lines 11 to 14 in algorithm \ref{algo:create-evo-chromosome}. This was done so that the patches remain as closely as possible within in the bounds of the image. Initially, the evolutionary algorithm can create patches which contain redundant pixels (for example the background, hair or clothing). We do not bias the algorithm towards initialising on patches of interest, such as the mouth or eyes.

\subsection{Mutation}

We propose two mutation operators to traverse through the search space. The first is a variation of the standard mutation operator and the second is tailored to the problem domain. When the mutation function is applied one of four things can be executed depending on the number of patches in the chromosomes, namely \textit{adding}, \textit{changing}, \textit{removing} or \textit{shifting}. The shift operation is a novel method which we describe below. Changing and shifting is always allowed, however adding and removing is only allowed under the following conditions:

\begin{itemize}
\item a patch can be added if the number of patches is less than the user-defined predefined maximum 
\item a patch can be removed if the number of patches is greater than the user-defined predefined minimum 
\end{itemize}

Adding a new patch simply generates a value for $x$ and $y$ in a similar way to the functions presented in algorithm \ref{algo:create-evo-chromosome}. The new patch is appended to the chromosome. Removing a patch consists of randomly selecting a $(x, y)$ pair to be removed. Changing a patch randomly selects one and then replaces the $(x, y)$ pair with new values.

\begin{figure}
  \centering
          \includegraphics[width=0.48\textwidth]{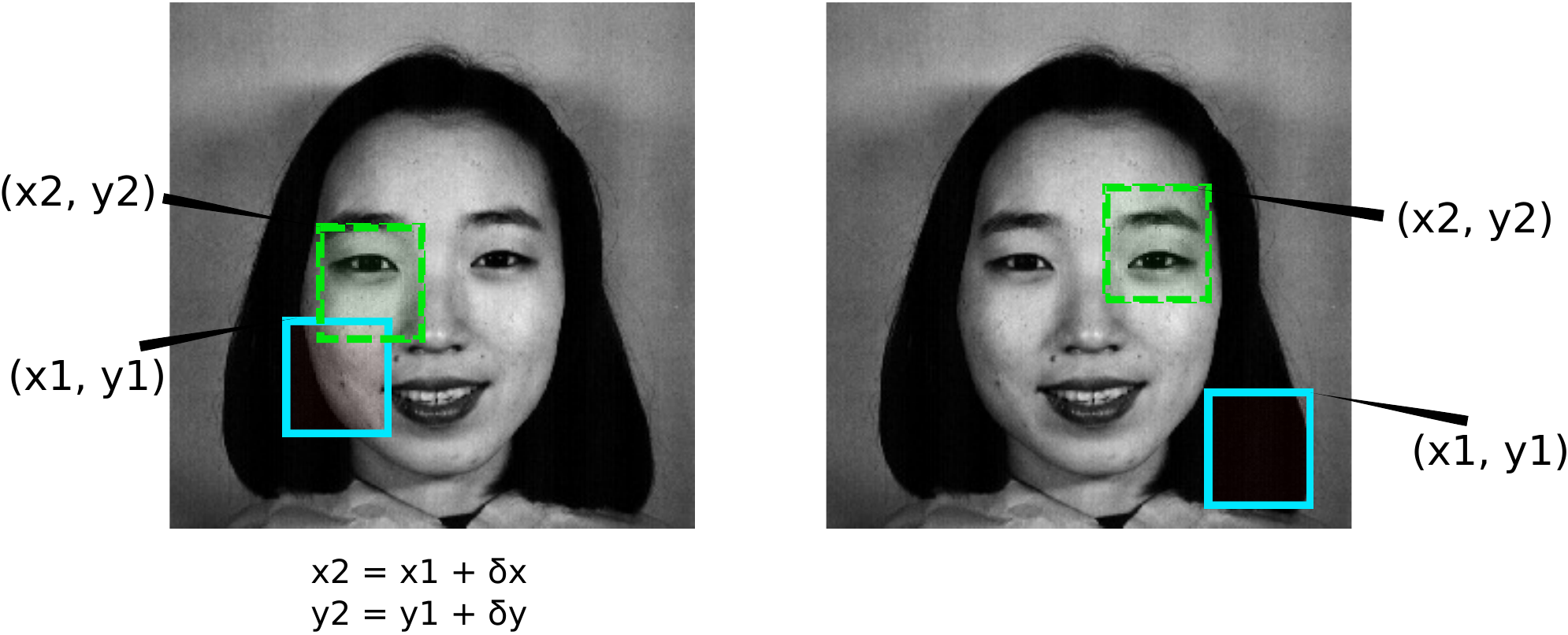}
  \caption[Illustrating shift and standard mutation]{Illustrating shift mutation on the left and standard mutation on the right. In shift mutation the patch is shifted, using values of $\delta$, within the associated bounds of $[-alpha$, $alpha]$ and $[-beta$, $beta]$. For the standard mutation the patch can jump anywhere. Expression image extracted from \cite{Lyons:1998:CodingFacial}} \label{fig:evofer-mutation}
\end{figure}

Algorithm \ref{algo:shift-mutation} presents the pseudocode for applying shift mutation to a chromosome. A patch is randomly selected from the chromosome and then the $x$ and $y$ values are extracted from the patch. The pair is then perturbed by values ranging between $[-alpha$, $alpha]$ and $[-beta$, $beta]$. This enables the algorithm to shift the patch in a random direction. The standard mutation operator allows the algorithm to take a large jump in the pixel space, whereas shift mutation restricts the jump. Figure \ref{fig:evofer-mutation} shows an example of the shift mutation and standard mutation operators being applied to a \textit{EvoFER} chromosome.

\begin{algorithm}

\SetKwData{mutatePatch}{mutation\_patch}
\SetKwData{alpha}{alpha}
\SetKwData{beta}{beta}
\SetKwData{chromosome}{chromosome}

\SetKwInOut{Input}{input}
\SetKwProg{Fn}{Function}{}{}

\Input{\alpha : patch width}
\Input{\beta : patch height}
\Input{\chromosome : the chromosome which will be mutated}

	\Begin{
	
	\mutatePatch $\leftarrow$ randomly select a patch from \chromosome

	$x_i$ $\leftarrow$ x value from \mutatePatch

	$y_i$ $\leftarrow$ y value from \mutatePatch

	$delta_x$ $\leftarrow$ $GenerateDeltaX(\alpha)$

	$delta_y$ $\leftarrow$ $GenerateDeltaY(\beta)$

	$x_i$ $\leftarrow$  $x_i + delta_x$

	$y_i$ $\leftarrow$  $y_i + delta_y$
	
\Return{\chromosome.}

\Fn{GenerateDeltaX (\alpha)}{
\Return random[$-alpha$, $alpha$]
}

\Fn{GenerateDeltaY (\beta)}{
\Return random[$-beta$, $beta$]
}

}
\caption{Applying shift mutation.}
\label{algo:shift-mutation} 
\end{algorithm}

\subsection{Crossover}

The standard crossover operator is applied to the $(x, y)$ pairs between two parent chromosomes. Specifically, let $(x_1, y_1)$ and $(x_2, y_2)$ be a pair of coordinates within parents $P_1$ and $P_2$. Then offspring $C_1$ is created by copying all of the genes in $P_1$ however, $x_1$ is replaced with $x_2$ and similarly, $y_1$ is replaced with $y_2$. Chromosome $C_2$ is created by copying all of the genes in $P_2$ however, $x_2$ is replaced with $x_1$ and similarly, $y_2$ is replaced with $y_1$. The offspring are evaluated and the one with the highest fitness is returned. 

\begin{figure} 
  \centering
          \includegraphics[width=0.48\textwidth]{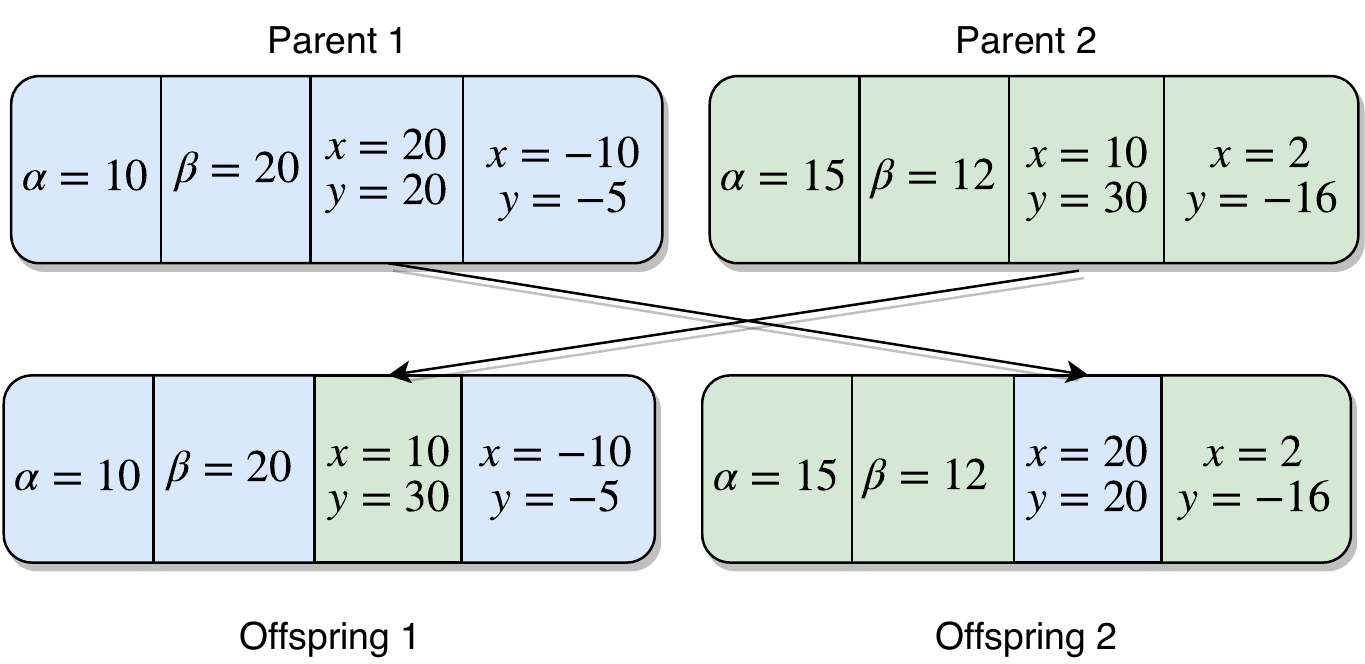}
  \caption[EvoFER crossover operator.]{Illustrating the crossover operator swapping the first patch between both parent chromosomes. All other genes within the parents are not swapped when creating the offspring.} \label{fig:evofer-crossover}
\end{figure}

Figure \ref{fig:evofer-crossover} shows an example of the crossover operator being applied to two \textit{EvoFER} chromosomes. The first patch from each parent chromosome is swapped to create the two offspring.

\subsection{Chromosome Evaluation}

Before executing the evolutionary process, we run a CNN model $M$ on the image dataset and record the validation accuracy and the number of neural network trainable parameters. Once the model is trained we apply the model to the test images and record the test performance -- we denote this as the baseline model.

Each chromosome $C_i$ is evaluated on every image $X_j$ in the dataset. When determining the fitness for $C_i$, the corresponding image patches $E_k$ are extracted from $X_j$ based on the number of $(x, y)$ pairs in $C_i$. All the extracted patches $E_k$ are stacked upon each other which produce a new image $N_j$. That is, each chromosome will be applied to each $X_j$ which creates a corresponding image $N_j$. The new images $N_j$ are input into the CNN model $M$. When computing the fitness score, we make use of the baseline validation accuracy and number of trainable CNN parameters. This is done to assign a fitness score that compares relative performance of chromosomes to the baseline CNN with the ultimate goal of improving the chromosome accuracy and reducing the number of parameters. 

We propose the fitness function presented in equation \ref{evofer-fitness}. The function considers both the effect of the validation accuracy and the number of trainable parameters between the network produced on images $N$ and the images $X$. The objective is to maximise the fitness. When the validation accuracy of the chromosome is larger than the baseline then $\frac{S_c}{S_b}$ is a large number. Similarly, when the number of parameters obtained by the network as a result of the chromosome is smaller than the number of parameters obtained by the baseline then $\frac{P_b - P_c}{P_b}$ is a large number. We allow a parameter $W_{S}$ to fine-tune the weight allocated to the validation accuracy. Small values (< 1.0) of $W_{S}$ will allocate greater importance to the number of parameters. This way, the experimenter can decide on the importance of the validation performance. The value of $W_{S}$ was determined through trial runs and we explored values ranging from \{0.1 to 5\}.

\begin{equation} \label{evofer-fitness}
\textnormal{Fitness (chromosome)} = \exp \left\{ W_{S} \big(\frac{S_c}{S_b}\big) + \big(\frac{P_b - P_c}{P_b}\big)\right\}
\end{equation}

where

\begin{conditions}
 \mbox{W\textsubscript{S}} &  weight of the validation accuracy \\
 \mbox{S\textsubscript{c}} &  validation accuracy for the chromosome \\
 \mbox{S\textsubscript{b}} &  validation accuracy for the baseline network \\
 \mbox{P\textsubscript{c}} & number of trainable weights for the chromosome network \\
 \mbox{P\textsubscript{b}} &  number of trainable weights for the baseline network \\
\end{conditions}

We provide an example of the fitness computation of a baseline and chromosome. Suppose that a CNN is trained on a FER dataset and the validation accuracy averaged over $R$ runs is computed to be 0.65. Assume that the number of trainable parameters for the network is 12,189,447. Now assume that the proposed algorithm is executed and that the validation accuracy of a chromosome is 0.31 and that the number of trainable parameters is 123,063 (a smaller value is obtained since the input images consist of smaller stacked patches in comparison to the larger original images). Then we have $\frac{0.31}{0.65} \approx 0.48$ and $\frac{12,189,447 - 123,063}{12,189,447} \approx 0.99$. Let $W_S = 5$. The final calculation for the fitness of the chromosome is $\exp{(5\times0.48 + 0.99)} \approx 29.67$. Since the objective is to maximise the fitness then a larger value denotes a better chromosome.

\begin{figure}
  \centering
          \includegraphics[width=0.48\textwidth]{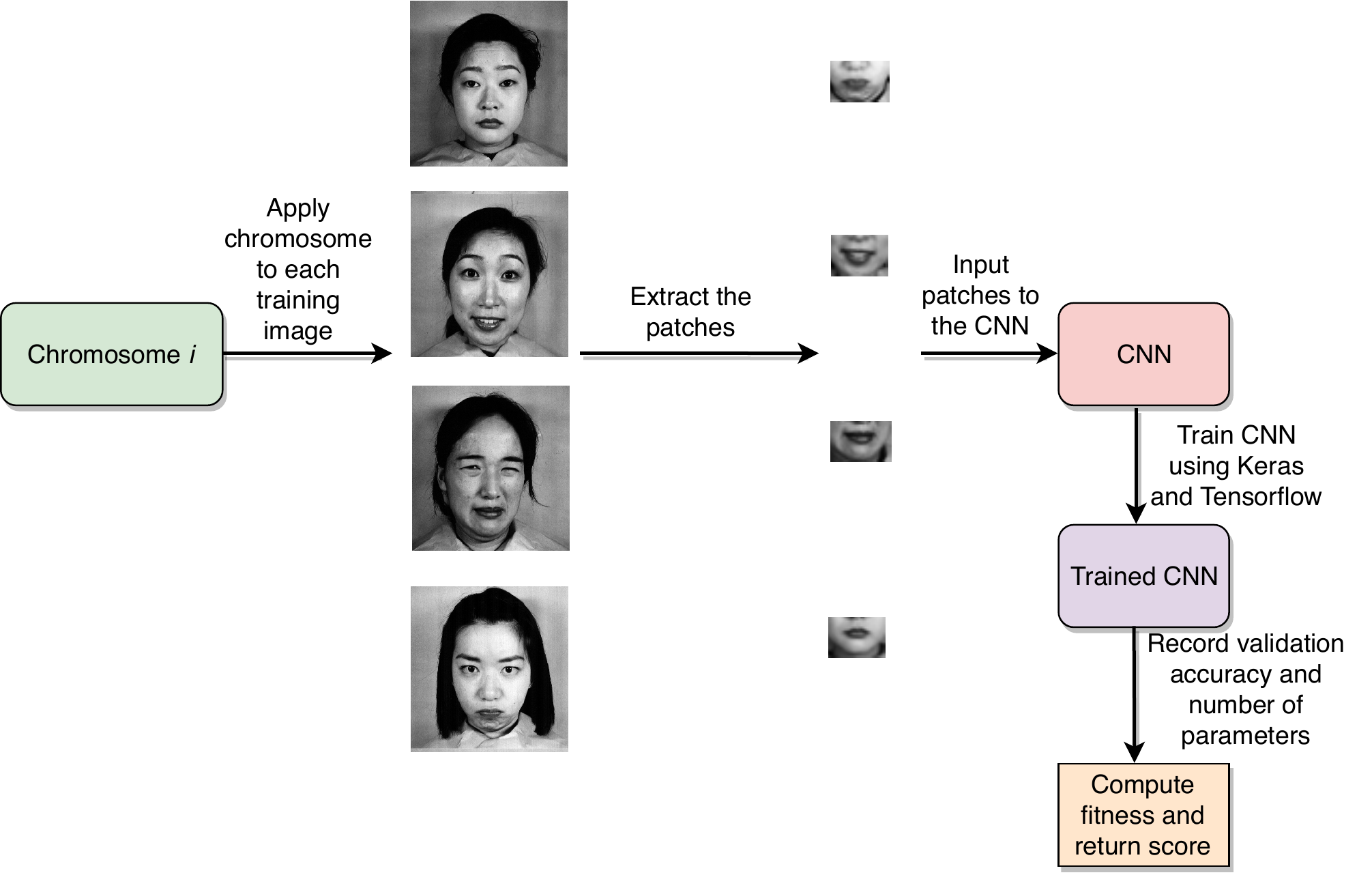}
  \caption[Illustrating the EvoFER pipeline.]{Illustrating the \textit{EvoFER} pipeline. Each chromosome is applied to the training data which in turns generates a new set of training images made up of stacked patches. The patches along with the associated target classes are input into the CNN.} \label{fig:evofer-pipeline}
\end{figure}

For each chromosome, we execute a CNN and allow it to train on the transformed images (made up of extracted patches). We use Keras and Tensorflow to train the CNN. The pipeline of constructing patches and training the CNN is illustrated in figure \ref{fig:evofer-pipeline}.

\section{Experimental Setup}\label{sec:evofer-setup}

In this section we describe the experimental setup which was used to evaluate the performance of \textit{EvoFER}. The rationale behind the decisions made to guide the experiments were based on preliminary implementations of CNNs to general FER.

\subsection{Datasets}

To evaluate the performance of \textit{EvoFER} we conducted a number of experiments on the datasets listed below. These datasets were selected since they are commonly used in the literature and represent different characteristics (gender, age, ethnic diversity and image resolution).

\begin{itemize}
\item JAFFE \cite{Lyons:1998:CodingFacial} 
\item KDEF \cite{KDEF:1998:KDEF} 
\item MUG \cite{TheMUU} 
\item RAFD \cite{langner2010presentation}
\end{itemize}

\subsection{Pre-processing}\label{sec:evofer-preprop}

Pre-processing is a vital step when dealing with image data. This can be described as some method to transform original input images using some image manipulation function. The function renders new images with the ultimate goal being that the new images will help the predictive performance of the classifier. Applying a CNN directly to images which have not been pre-processed in some way can yield weaker predictive performance. For example, consider a dataset containing images of people for which the lighting conditions across the face are not consistent. It would be more suitable to attempt to correct the variation in illumination in such a way to have the same amount of lighting exposure across the entire face so that the CNN can learn useful features across the entire face. Pre-processing was shown to improve classification performance in a number of studies \cite{Yu:2015:ImageBased, Kim:2016:FusingAligned, Shin:2016:BaselineCNN, Spiers:2016:FacialEmotion, Fasel:2002:HeadPose, tang:2013:DeepLearningUsing, Sang:2017:FERUsing}. 

We implemented histogram equalisation on each of the images. Additionally, we converted each image into greyscale as the additional information available in colour images does not impact the performance of FER. To achieve this, we made use of OpenCV as it is commonly used in literature, for examples see \cite{Xu:2015:FERBasedOn, Jung:2015:DevelopmentOf, Zavarez:2017:CrossDatabase}.

\subsection{Data Augmentation}

The application of CNNs often require large datasets to enable good predictive performance and has been successfully used in literature, for examples see \cite{Valero:2016:AutomaticFER, Zhang:2016:ADeepNN, Alizadeh:2017:CNNForFER, Sang:2017:FERUsing,  Kim:2016:FusingAligned}. We implemented a number of augmentation techniques using the literature to guide our decisions. We augment our training images using the following techniques:

\begin{itemize}
\item horizontal flipping
\item blurring
\item noise
\item rotate by -5 and -10 degrees
\item rotate by 5 and 10 degrees
\end{itemize}

\begin{figure}
  \centering
          \includegraphics[width=0.48\textwidth]{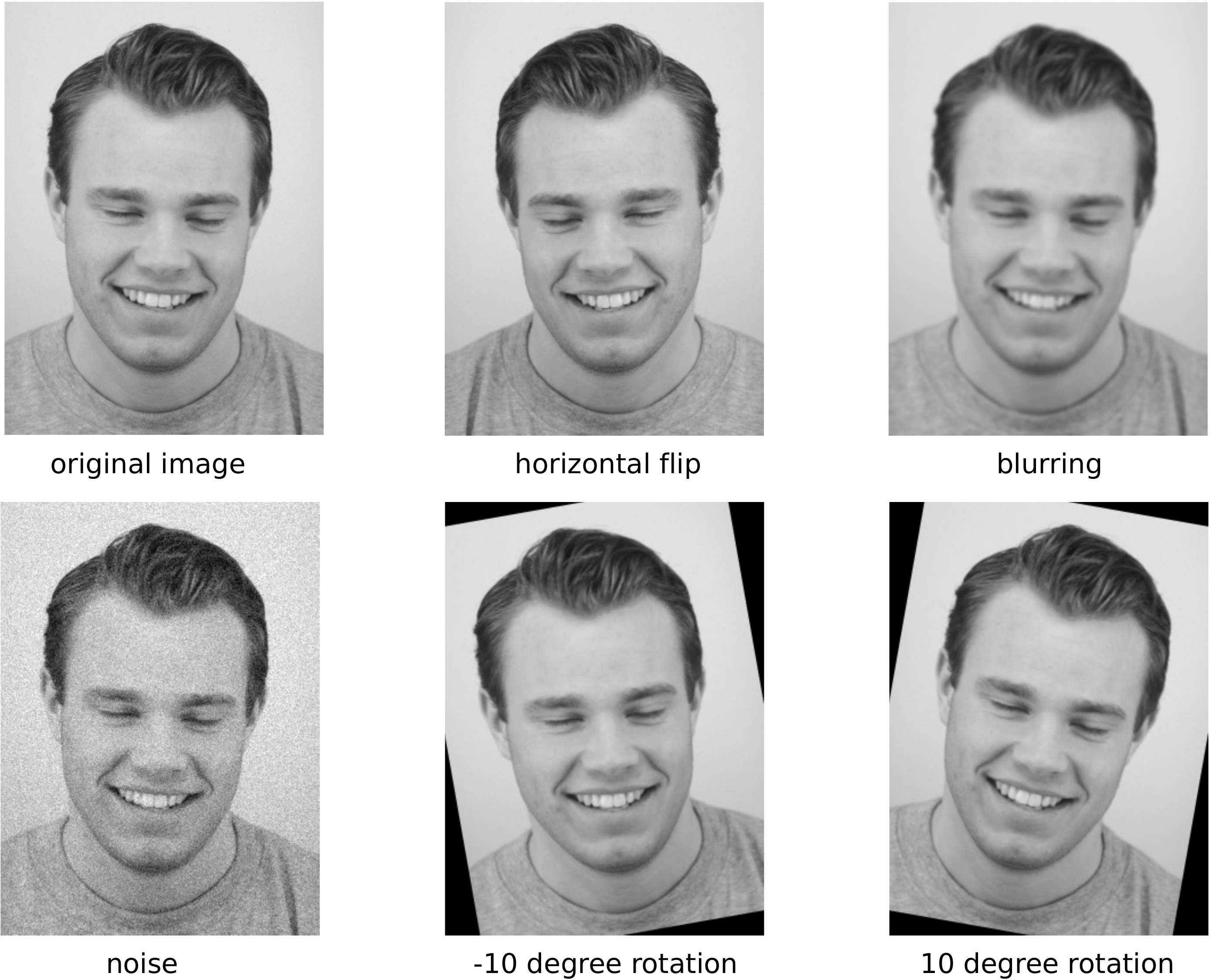}
  \caption[Different augmentation techniques which are used in \textit{EvoFER}.]{Illustrating the different augmentation techniques which were used to increase the number of training images. Images extracted from \cite{KDEF:1998:KDEF}.} \label{fig:evofer-augment}
\end{figure}

These augmentation techniques were only applied to the training images. The test images were kept in their original form. For each training image we generated an additional 9 images by applying two levels of blurring and noise. These were achieved by using \textit{imgaug}\footnote{https://github.com/aleju/imgaug} -- a software package for image augmentation. Table \ref{table:augmented-images} presents the number of images which were used for training for each dataset after we applied the augmentation techniques. Figure \ref{fig:evofer-augment} presents the augmentation techniques applied to each image.

\begin{center}
\begin{table}[!h]
\begin{centering}
\caption[Number of training images used in the \textit{EvoFER} study for each dataset after the images were augmented. ]{Number of training images used for each dataset after the images were augmented. The images were augmented using rotation, blurring, random noise and horizontal flipping.}\label{table:augmented-images}
\begin{tabular}{cc}

\hline 
\textbf{Dataset} & \textbf{Training images after augmentation}\tabularnewline
\hline 
JAFFE & 1,910\tabularnewline
KDEF & 6,860\tabularnewline
RAFD & 10,152\tabularnewline
MUG & 2,800\tabularnewline
\hline 
\end{tabular}
\par\end{centering}

\end{table} 
\par\end{center}

\subsection{Network Architecture}

\begin{figure}
  \centering
          \includegraphics[width=0.48\textwidth]{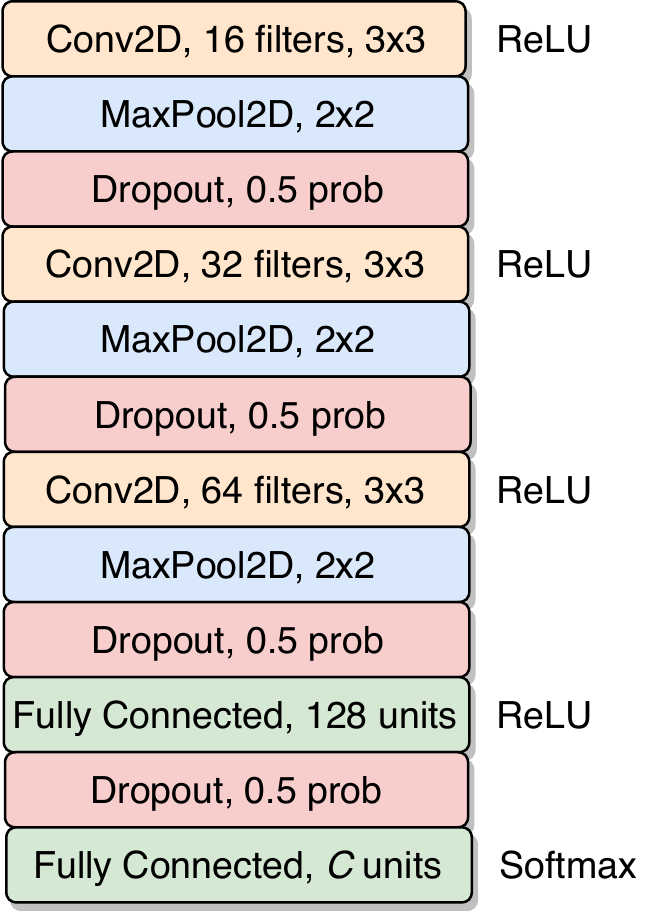}
  \caption[CNN architecture used in EvoFER experiment.]{The CNN architecture which is proposed for the experiments. This network was inspired by the findings from existing literature. 'Conv2D' denotes 2D convolution and 'MaxPool2D' denotes 2D max pooling. The associated parameters are listed.} \label{fig:evofer-network}
\end{figure}

Figure \ref{fig:evofer-network} presents the CNN architecture which we propose to use for the \textit{EvoFER} experiments. The architecture was inspired by our findings from existing literature, for examples see \cite{Alizadeh:2017:CNNForFER, Sang:2017:FERUsing, Barsoum:2016:TrainingDeep, Raghuvanshi:2016:FER}. More specifically, the reviewed works used on average 3 convolutional layers with max pooling in between them, and two fully connected layers at the end of the network. The ReLU activation function was used the most frequently in all layers except for the last where softmax was used. We thus used the literature to guide our decisions, and furthermore various modifications of the architecture were explored using preliminary runs on the JAFFE dataset by varying the depth, activation function and parameters. The last fully connected layer takes on a value of `C' for the number of units as this is dependant on the number of expression classes in each dataset.

The proposed architecture consists of three convolutional and max pooling layers, along with two fully connected layers. Dropout was applied after each of the layers except for the last one. The ReLU activation function was used for all the layers except for the last fully connected layer whereby the softmax function was used. It would also be possible to optimise the network using machine learning, such as  \textit{EDEN}, as described in \cite{Dufourq:2017:EDEN}, however, we chose to use the findings from the literature to guide our decisions. Studies in literature (for examples see \cite{Zhou:2016:FERBased, AlShabi:2016:FER, Rassadin:2017:GroupLevelER}) reveal that using an ensemble will result in superior classification performance, however, we wanted to examine the effect of using patches as input to enhance the performance. 

We ran two sets of experiments. In the first, we ran \textit{EvoFER} using the literature inspired network and used the patches as input to the CNN. In the second set of experiments, we ran the same literature inspired network and used the full image as input to the CNN. We denote experiments conducted on the full images as the \textit{baseline}.

\subsection{Training and testing}

Lopes \textit{et al.} \cite{Lopes:2017:FER} describe that a fair comparison is one whereby the same subjects should not be found in the training and testing sets. In their work they split the images into 8 groups containing a number of subjects each, and that the same subject is not found in more than one group. From all the studies reviewed, it was found that their work contained the most emphasis on fairness. We implemented this approach in our experiments, for which 30\% of the data was used for testing and the remaining for training. We repeated the execution of \textit{EvoFER} ten times and averaged our results. This was conducted for both the evaluation of the baseline and for the evolutionary process. The choice of optimiser was selected based on a literature survey. We selected the Adam optimiser and a batch size of 8.

\subsection{\textit{EvoFER} Parameters}

Table \ref{table-params-ea} provides the details of the parameters associated with the evolutionary algorithm. The parameters associated with the training of the CNN are presented in table \ref{table-params-cnn-evofer}. Finally, table \ref{table-params-init-evofer} presents the parameters associated with the initialisation of the chromosomes during the initial population generation and mutation operation. These parameters were chosen by performing a search on a number of values which were found in the literature and by performing a random hyper-parameter search.

\begin{table}[h]
\begin{centering}
\caption[Parameters associated with the evolutionary algorithm in the \textit{EvoFER} study.]{Parameters associated with the evolutionary algorithm.}
\label{table-params-ea}
\begin{tabular}{cc}

\hline 
\textbf{Parameter} & \textbf{Value}\tabularnewline
\hline 
Population size & 100\tabularnewline
Tournament size & 7\tabularnewline
Crossover percentage & 50\tabularnewline
Mutation percentage & 50\tabularnewline
Generations & 15\tabularnewline
Fitness function, $W_S$ & 5\tabularnewline
\hline 
\end{tabular}
\par\end{centering} 

\end{table}

\begin{table}[h]
\begin{centering}
\caption[Parameters associated with the CNN in the \textit{EvoFER} study.]{Parameters associated with the CNN.}\label{table-params-cnn-evofer}
\begin{tabular}{cc}

\hline 
\textbf{Parameter} & \textbf{Value}\tabularnewline
\hline 
Number of epochs & 10\tabularnewline
Optimiser & Adam\tabularnewline
Batch size & 8\tabularnewline
\hline 
\end{tabular}
\par\end{centering}

\end{table}

\begin{table}
\begin{centering}
\caption[Additional parameters used in the \textit{EvoFER} study.]{Additional parameters used in the initial population generation and
mutation operator.}\label{table-params-init-evofer}
\begin{tabular}{cc}

\hline 
\textbf{Parameter} & \textbf{Value}\tabularnewline
\hline 
min\_alpha & 30\tabularnewline
max\_alpha & 50\tabularnewline
min\_beta & 30\tabularnewline
max\_beta & 50\tabularnewline
min\_patches & 1\tabularnewline
max\_patches & 4\tabularnewline
\hline 
\end{tabular}
\par\end{centering}

\end{table}

\section{Results and Discussion}\label{sec:evofer-results}

Table \ref{table:evofer-accuracy} presents the average test classification accuracy results. The baseline accuracy (literature inspired CNN model with original images) and \textit{EvoFER} accuracy (the same literature inspired CNN model with extracted patches from best chromosome) are presented. The CNN architecture and hyper-parameters were the same for both results. The findings reveal that on all of the datasets \textit{EvoFER} was able to outperform the same CNN model which had been trained on the original images. The smallest improvement in accuracy was observed in the KDEF dataset with an improvement of 1.3\%. The largest improvement in classification accuracy was obtained on the MUG dataset with a value of 20.6\%. This indicates that given a fixed architecture, an EA can be applied in such a way to extract patches from an image and to train the CNN on those patches and achieve competitive performance.

\begin{table}
\begin{centering}

\caption[Test classification accuracy (\%) for the CNN network on the original
images and performance when using \textit{EvoFER}.]{Test classification accuracy (\%) for the CNN network on the original images and performance when using \textit{EvoFER}. The standard deviation is presented in parentheses. The findings reveal that the performance is superior when \textit{EvoFER} is used.}\label{table:evofer-accuracy}

\begin{tabular}{>{\centering}p{1.5cm}>{\centering}p{1.5cm}>{\centering}p{1.5cm}>{\centering}p{1.5cm}}
\hline 
\textbf{Dataset} & \textbf{Baseline Accuracy} & \textbf{EvoFER Accuracy} & \textbf{Difference}\tabularnewline
\hline 
KDEF & 61.9  (1.6 )& 63.2 (2.9) & 1.3\tabularnewline
JAFFE & 60.0  (4.7) & 75.5 (4.4) & 15.5\tabularnewline
RAFD & 75.4 (3.4)  & 80.4 (2.3 )& 5.0\tabularnewline
MUG & 45.8 (3.8) & 66.4 (5.0) & 20.6\tabularnewline
\hline 
\end{tabular}
\par\end{centering} 
\end{table}

Why did \textit{EvoFER} achieve superior performance to the baseline method? It is hypothesised that \textit{EvoFER} is able to obtain better predictive performance given that it inputs extracted patches to the CNN and consequently the number of neural network parameters are reduced. The average number of neural network trainable parameters obtained in the experiments are presented in table \ref{table:evofer-parameters}. For each dataset the number of neural network parameters for \textit{EvoFER} is significantly less than for the baseline CNN. The baseline CNN inputs larger images of the faces compared to \textit{EvoFER} which inputs smaller extracted patches. The percentage difference achieved by \textit{EvoFER} for KDEF, JAFFE, RAFD and MUG are 88\%, 97\%, 97\% and 98\% respectively. The largest difference in parameters was obtained on the JAFFE dataset with a reduction of over 7 million parameters. It is observed on the JAFFE dataset that the standard deviation for the number of parameters is zero, which indicates that in each execution of the algorithm, \textit{EvoFER} extracted the same number of patches of the same size. Despite the large standard deviation on the KDEF dataset the number of parameters is still distant from the baseline parameters.

\begin{figure} 
  \centering
          \includegraphics[width=0.48\textwidth]{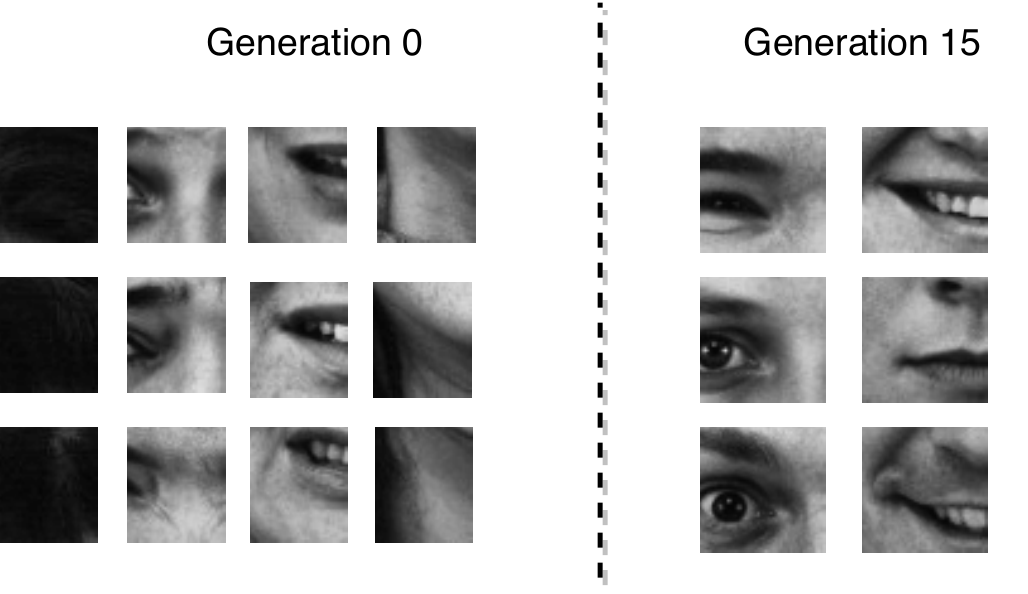}
  \caption[Patches extracted from the best chromosome on KDEF dataset.]{Illustrating the patches extracted from the best chromosome in generation 0 and in the last evolutionary generation. Each row presents the patches which were extracted for randomly selected test examples in the KDEF dataset.} \label{fig:evofer-patches-kdef}
\end{figure}

\begin{figure} 
  \centering
          \includegraphics[width=0.48\textwidth]{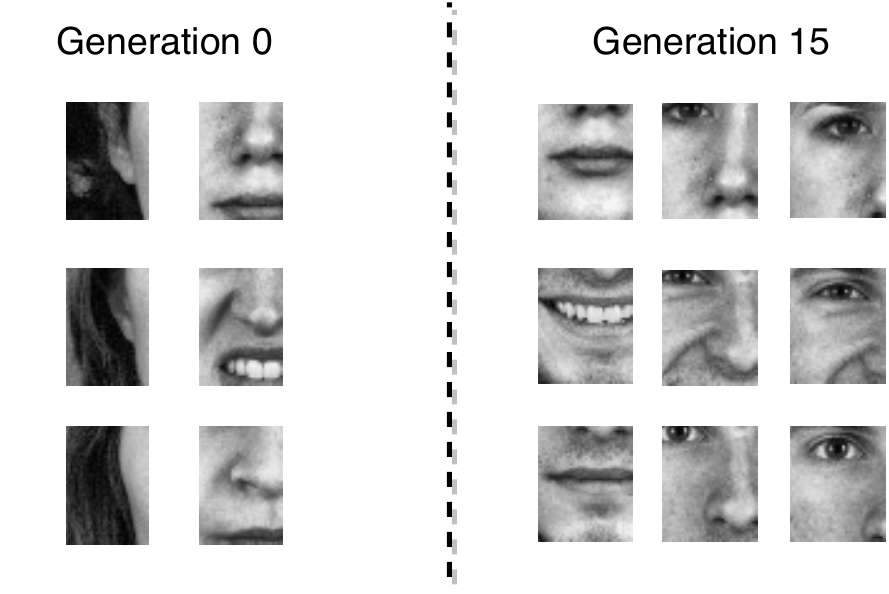}
  \caption[Patches extracted from the best chromosome on MUG dataset.]{Illustrating the patches extracted from the best chromosome in generation 0 and in the last evolutionary generation. Each row presents the patches which were extracted for randomly selected test examples in the MUG dataset.} \label{fig:evofer-patches-mug}
\end{figure}

\begin{figure}
  \centering
          \includegraphics[width=0.48\textwidth]{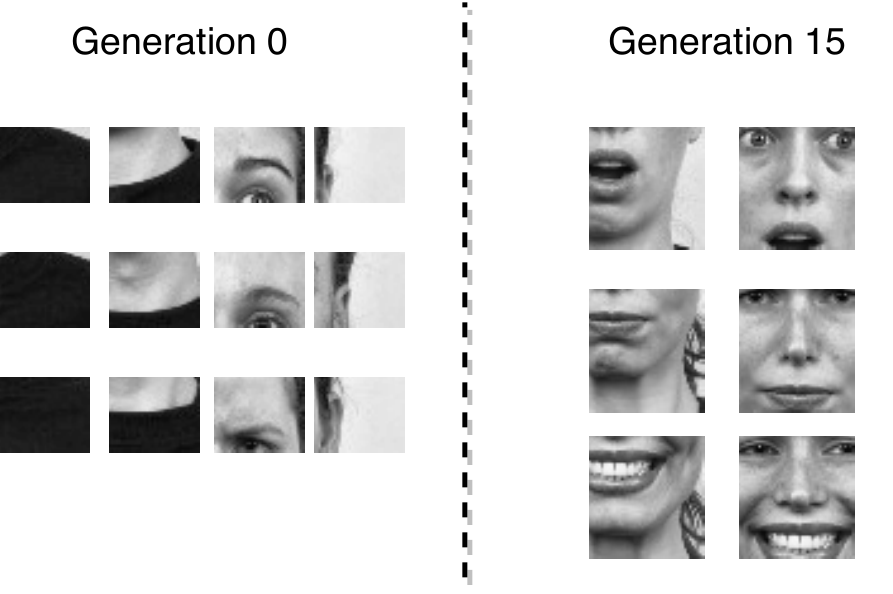}
  \caption[Patches extracted from the best chromosome on RAFD dataset.]{Illustrating the patches extracted from the best chromosome in generation 0 and in the last evolutionary generation. Each row presents the patches which were extracted for randomly selected test examples in the RAFD dataset.} \label{fig:evofer-patches-rafd}
\end{figure}

Figures \ref{fig:evofer-patches-kdef} to \ref{fig:evofer-patches-rafd} illustrate the patches which were extracted from the best chromosome in generation 0 and the last generation for the various datasets. Each row presents the patches which were extracted from a random test example. 

\begin{table}
\begin{centering}

\caption[The average number of trainable neural network parameters when using
the original image and \textit{EvoFER} extracted patches.]{The average number of trainable neural network parameters when using
the original image and \textit{EvoFER} extracted patches. The standard deviation is presented in parentheses.}\label{table:evofer-parameters}

\begin{tabular}{>{\centering}p{1.0cm}>{\centering}p{1.5cm}>{\centering}p{2.0cm}>{\centering}p{1.5cm}>{\centering}p{0.6cm}}
\hline 
\textbf{Dataset} & \textbf{Baseline Parameters} & \textbf{EvoFER Parameters} & \textbf{Difference} & \textbf{\%} \tabularnewline
\hline 
KDEF & 1,218,944 & 145,741 (15,782) & 1,073,203 & 88.0\tabularnewline
JAFFE & 7,397,127 & 155,543 (0) & 7,241,584 & 97.9\tabularnewline
RAFD & 2,152,520 & 45,528 (6,476) & 2,106,992 & 97.9\tabularnewline
MUG & 2,574,279 & 47,027 (6,345) & 2,527,252 & 98.2\tabularnewline
\hline 
\end{tabular}
\par\end{centering} 
\end{table}

For the JAFFE dataset, the best chromosome from the initial and final population extracted two patches. For generation 0, both patches were around the left eye region. This of course is sub-optimal as that does not provide the CNN with enough information to discriminate between the various emotion classes. The best chromosome extracts patches around the left eye region and includes the eyebrow. The second patch extracts pixels around the center and right region of the mouth. 

Figure \ref{fig:evofer-patches-kdef} illustrates the patches for the KDEF dataset. In generation 0 the chromosome extracts four patches. The first and last patch are redundant as the first extracts pixels around the hair and the other around the neck. This does not help discriminate between expressions. The second and third patch are better since they extract pixels between the eyes and near the left corner of the mouth. In the case of the best chromosome in generation 15, two patches are extracted. Combined, they extract pixels near the left eye and half of the mouth.

The extracted patches for the MUG dataset are presented in figure \ref{fig:evofer-patches-mug}. The best chromosome from generation 0 extracts two patches, one near the left ear and the other contains pixels around the nose and mouth. The first patch does not assist in distinguishing between expressions. The best chromosome from the final generation extracts three patches. The second and the third patch are similar since they both extract pixels around the left eye. The first patch extracts pixels around the center of the mouth.

Finally, figure \ref{fig:evofer-patches-rafd} presents the patches extracts for the RAFD dataset. For generation 0, the best chromosome extracts four patches of which the first, second and last contain redundant information. Those patches cannot help the CNN distinguish between the various expression classes. The third patch extracts pixels near the right eye and eyebrow. For the last generation, the best chromosome extracts two patches. The first contains pixels around the mouth and chin, and the second contains pixels capturing the mouth, nose and eyes (no information about the eyebrows).

From the figures it is observed that the best predictive performance is achieved when pixels around the eyes and mouth are obtained. The best chromosome from each of the last generations extracted two patches on 3 of the 4 datasets. We can deduce that those areas are the most salient when distinguishing between expressions on the datasets examined. We did not have sufficient computing resources to compare the results to state-of-the-art CNNs which use much deeper network with a larger number of parameters. The findings do reveal that on a fixed architecture which was inspired by the literature, \textit{EvoFER} is able to enhance the performance by using stacked patches as input.

What is the processing time for \textit{EvoFER} given that the nose has to be located prior to the application of the chromosome? We examined the time it took to perform the necessary pre-processing steps for \textit{EvoFER} and in the case of the baseline CNN. These findings are presented in table \ref{table-evofer-times}. The time, in seconds, is presented for the two methods to pre-process and predict the expression for 10 images. In the case of KDEF, RAFD and MUG, \textit{EvoFER} took approximately twice as long as the baseline method. This is because \textit{EvoFER} has the extra overhead of needing to locate the nose to establish a reference point for the chromosome. \textit{EvoFER} took less time than the baseline on the JAFFE dataset. It can be hypothesised that this is the case since the resolution of the images in the JAFFE dataset is much smaller than the other datasets. It would be of interest to reduce the resolution of images in the other datasets to determine if this could result in faster execution times for \textit{EvoFER}. Despite the fact that \textit{EvoFER} has an additional overhead, the processing time is not significantly large and could thus still be implemented in a real-world setting. In terms of training, \textit{EvoFER} took up to 5 hours whereas training the equivalent network in a traditional setting (i.e. the baseline) took less than a minute. Once the location of the patches have been optimised, \textit{EvoFER} trains faster than the baseline approach (i.e. the same network on the original dataset) - these findings are presented in table \ref{table:train-time}. 

\begin{table}
\caption{A comparison between the training time (seconds) when using the baseline network on the original input and using \textit{EvoFER} on the extracted patches input. In both cases the same network architecture and hyper-parameters are used. The primary difference is the input.}\label{table:train-time}
\begin{tabular}{c|>{\centering}p{3cm}|>{\centering}p{3cm}}
\hline 
\textbf{Dataset} & \textbf{Baseline with original dataset input} & \textbf{EvoFER with extracted patches input}\tabularnewline
\hline 
KDEF & 247 & 42\tabularnewline
\hline 
JAFFE & 61 & 10\tabularnewline
\hline 
RAFD & 112 & 30\tabularnewline
\hline 
MUG & 43 & 10\tabularnewline
\hline 
\end{tabular}
\end{table}

\begin{table}
\begin{centering}
\caption[Average time taken (seconds) to process 10 images using \textit{EvoFER} and the baseline CNN architecture.]{Average time taken (seconds) to process 10 images using \textit{EvoFER} and the baseline. The results are averaged over 10 executions. The standard deviation is shown in parentheses.}\label{table-evofer-times}
\begin{tabular}{ccc}
\hline 
\textbf{Dataset} & \textbf{EvoFER} & \textbf{Baseline}\tabularnewline
\hline 
JAFFE & 0.27 (0.01) & 0.29 (0.01)\tabularnewline
KDEF & 0.42 (0.01) & 0.18 (0.01)\tabularnewline
RAFD & 0.28 (0.01)& 0.14 (0.01)\tabularnewline
MUG & 0.23 (0.01)& 0.13 (0.01)\tabularnewline
\hline 
\end{tabular}
\par\end{centering}  
\end{table}

\section{Conclusion}\label{sec:evofer-conclusion}

This study proposes a novel evolutionary algorithm to extract patches from images with the goal of transforming images into more compact representations whilst retaining the predictive performance (or increasing it) and to reduce the number of trainable parameters. We propose a chromosome representation which allows the algorithm to encode locations relative to the nose. The chromosomes also encode the width and height of the patches to be extracted at each location. We introduce a fitness function which combines the relative performance of each chromosome to a baseline execution. The baseline execution consists of running a CNN on the original images. The multi-objective fitness function attempts to optimise the validation accuracy and number of trainable parameters. To enable us to interface each chromosome to the training of a CNN we use Keras and Tensorflow. 

We evaluated \textit{EvoFER} and the findings revealed that it can achieve superior performance to the exact CNN architecture trained on the entire image.  Furthermore, the findings revealed that \textit{EvoFER} can reduce the number of trainable parameters, on average, up to 95\%. \textit{EvoFER} was able to optimise the search for optimal patch locations and size to enable the CNN to distinguish between the various expressions (without any insights hardcoded into the algorithm). Initially \textit{EvoFER} could select patches which did not contain parts of the face at all. In several cases there was patches which were located near people's hair. Through the evolutionary process, \textit{EvoFER} extracted patches generally around the eyes and mouth which enabled the CNN to achieve better predictive performance. We hypothesise that superior performance could be achieved by increasing the number of epochs; in this study we imposed a limit due to available computational time. 

\section*{Acknowledgment}
Part of this work was included in the PhD thesis titled `Evolutionary Deep Learning' by Emmanuel Dufourq and was accepted in 2019 by the University of Cape Town. The financial assistance of the National Research Foundation (NRF) towards this research is hereby acknowledged. Opinions expressed and conclusions arrived at, are those of the authors and are not necessarily to be attributed to the NRF.

\bibliographystyle{IEEEtran}

\begin{thebibliography}{10}
\providecommand{\url}[1]{#1}
\csname url@samestyle\endcsname
\providecommand{\newblock}{\relax}
\providecommand{\bibinfo}[2]{#2}
\providecommand{\BIBentrySTDinterwordspacing}{\spaceskip=0pt\relax}
\providecommand{\BIBentryALTinterwordstretchfactor}{4}
\providecommand{\BIBentryALTinterwordspacing}{\spaceskip=\fontdimen2\font plus
\BIBentryALTinterwordstretchfactor\fontdimen3\font minus
  \fontdimen4\font\relax}
\providecommand{\BIBforeignlanguage}[2]{{%
\expandafter\ifx\csname l@#1\endcsname\relax
\typeout{** WARNING: IEEEtran.bst: No hyphenation pattern has been}%
\typeout{** loaded for the language `#1'. Using the pattern for}%
\typeout{** the default language instead.}%
\else
\language=\csname l@#1\endcsname
\fi
#2}}
\providecommand{\BIBdecl}{\relax}
\BIBdecl

\bibitem{Tian:2011:fFER}
Y.~Tian, T.~Kanade, and J.~F. Cohn, ``Facial expression recognition,'' in
  \emph{Handbook of face recognition}.\hskip 1em plus 0.5em minus 0.4em\relax
  Springer, 2011, pp. 487--519.

\bibitem{Jack:2015:TheHuman}
R.~Jack and P.~Schyns, ``The human face as a dynamic tool for social
  communication,'' \emph{Current Biology}, vol.~25, no.~14, pp. R621 -- R634,
  2015.

\bibitem{Calenau:2013:FERABrief}
C.~D. C?leanu, ``Face expression recognition: A brief overview of the last
  decade,'' in \emph{2013 IEEE 8th International Symposium on Applied
  Computational Intelligence and Informatics}, May 2013, pp. 157--161.

\bibitem{LeCun:2015:DeepLearning}
Y.~LeCun, Y.~Bengio, and G.~Hinton, ``Deep learning,'' \emph{Nature}, vol. 521,
  no. 7553, p. 436, 2015.

\bibitem{LeCun:1990:HandwrittenDigit}
Y.~LeCun, B.~E. Boser, J.~S. Denker, D.~Henderson, R.~E. Howard, W.~E. Hubbard,
  and L.~D. Jackel, ``Handwritten digit recognition with a back-propagation
  network,'' in \emph{{Advances in Neural Information Processing Systems}},
  1990, pp. 396--404.

\bibitem{Corneanu:2016:SurveyonRGB}
C.~A. Corneanu, M.~O. Simon, J.~F. Cohn, and S.~E. Guerrero, ``Survey on rgb,
  3d, thermal, and multimodal approaches for facial expression recognition:
  History, trends, and affect-related applications,'' \emph{IEEE Transactions
  on Pattern Analysis and Machine Intelligence}, vol.~38, no.~8, pp.
  1548--1568, Aug 2016.

\bibitem{pramerdorfer:2016:FERUsing}
C.~Pramerdorfer and M.~Kampel, ``Facial expression recognition using
  convolutional neural networks: State of the art,'' \emph{arXiv preprint
  arXiv:1612.02903}, 2016.

\bibitem{Pantic:2000:AutomaticAnalysis}
M.~Pantic and L.~J.~M. Rothkrantz, ``Automatic analysis of facial expressions:
  the state of the art,'' \emph{IEEE Transactions on Pattern Analysis and
  Machine Intelligence}, vol.~22, no.~12, pp. 1424--1445, Dec 2000.

\bibitem{Sariyanidi:2015:AutomaticAnalysis}
E.~Sariyanidi, H.~Gunes, and A.~Cavallaro, ``Automatic analysis of facial
  affect: A survey of registration, representation, and recognition,''
  \emph{IEEE Transactions on Pattern Analysis and Machine Intelligence},
  vol.~37, no.~6, pp. 1113--1133, June 2015.

\bibitem{Wu:2012:SurveyOf}
T.~Wu, S.~Fu, and G.~Yang, ``Survey of the facial expression recognition
  research,'' in \emph{Advances in Brain Inspired Cognitive Systems}, H.~Zhang,
  A.~Hussain, D.~Liu, and Z.~Wang, Eds.\hskip 1em plus 0.5em minus 0.4em\relax
  Berlin, Heidelberg: Springer Berlin Heidelberg, 2012, pp. 392--402.

\bibitem{Martinez:2016:AdvancesChallenges}
B.~Martinez and M.~F. Valstar, \emph{Advances, Challenges, and Opportunities in
  Automatic Facial Expression Recognition}.\hskip 1em plus 0.5em minus
  0.4em\relax Cham: Springer International Publishing, 2016, pp. 63--100.

\bibitem{ghayoumi:2017:AQuickReview}
M.~Ghayoumi, ``A quick review of deep learning in facial expression,''
  \emph{Journal of Communication and Computer}, vol.~14, pp. 34--38, 2017.

\bibitem{Zhang:2018:FERBased}
T.~Zhang, \emph{Facial Expression Recognition Based on Deep Learning: A
  Survey}.\hskip 1em plus 0.5em minus 0.4em\relax Cham: Springer International
  Publishing, 2018, pp. 345--352.

\bibitem{Ko:2018:ABriefReview}
B.~C. Ko, ``A brief review of facial emotion recognition based on visual
  information,'' \emph{Sensors}, vol.~18, no.~2, p. 401, 2018.

\bibitem{Latha:2016:AReviewOnDeep}
C.~P. Latha and M.~Priya, ``A review on deep learning algorithms for speech and
  facial emotion recognition,'' \emph{APTIKOM Journal on Computer Science and
  Information Technologies}, vol.~1, no.~3, pp. 88--104, 2016.

\bibitem{Miller:1996:GAS:1326713.1326715}
B.~L. Miller and D.~E. Goldberg, ``Genetic algorithms, selection schemes, and
  the varying effects of noise,'' \emph{Evolutionary Computation}, vol.~4,
  no.~2, pp. 113--131, Jun. 1996.

\bibitem{KDEF:1998:KDEF}
D.~Lundqvist, F.~A., and A.~Ohman, ``The karolinska directed emotional faces
  ? kdef,'' 1998.

\bibitem{Lyons:1998:CodingFacial}
M.~Lyons, S.~Akamatsu, M.~Kamachi, and J.~Gyoba, ``Coding facial expressions
  with gabor wavelets,'' in \emph{Proceedings Third IEEE International
  Conference on Automatic Face and Gesture Recognition}, Apr 1998, pp.
  200--205.

\bibitem{TheMUU}
N.~Aifanti, C.~Papachristou, and A.~Delopoulos, ``The mug facial expression
  database,'' in \emph{11th International Workshop on Image Analysis for
  Multimedia Interactive Services}, April 2010, pp. 1--4.

\bibitem{langner2010presentation}
O.~Langner, R.~Dotsch, G.~Bijlstra, D.~H. Wigboldus, S.~T. Hawk, and
  A.~Van~Knippenberg, ``Presentation and validation of the radboud faces
  database,'' \emph{Cognition and emotion}, vol.~24, no.~8, pp. 1377--1388,
  2010.

\bibitem{Yu:2015:ImageBased}
Z.~Yu and C.~Zhang, ``Image based static facial expression recognition with
  multiple deep network learning,'' in \emph{Proceedings of the 2015 ACM on
  International Conference on Multimodal Interaction}, ser. International
  Conference on Multimodal Interaction '15.\hskip 1em plus 0.5em minus
  0.4em\relax New York, NY, USA: ACM, 2015, pp. 435--442.

\bibitem{Kim:2016:FusingAligned}
B.~K. Kim, S.~Y. Dong, J.~Roh, G.~Kim, and S.~Y. Lee, ``Fusing aligned and
  non-aligned face information for automatic affect recognition in the wild: A
  deep learning approach,'' in \emph{2016 IEEE Conference on Computer Vision
  and Pattern Recognition Workshops}, June 2016, pp. 1499--1508.

\bibitem{Shin:2016:BaselineCNN}
M.~Shin, M.~Kim, and D.~S. Kwon, ``Baseline cnn structure analysis for facial
  expression recognition,'' in \emph{25th IEEE International Symposium on Robot
  and Human Interactive Communication}, Aug 2016, pp. 724--729.

\bibitem{Spiers:2016:FacialEmotion}
D.~Spiers, ``Facial emotion detection using deep learning,'' 2016.

\bibitem{Fasel:2002:HeadPose}
B.~Fasel, ``Head-pose invariant facial expression recognition using
  convolutional neural networks,'' in \emph{Proceedings. Fourth IEEE
  International Conference on Multimodal Interfaces}, 2002, pp. 529--534.

\bibitem{tang:2013:DeepLearningUsing}
Y.~Tang, ``Deep learning using linear support vector machines,'' \emph{arXiv
  preprint arXiv:1306.0239}, 2013.

\bibitem{Sang:2017:FERUsing}
D.~V. Sang, N.~V. Dat, and D.~P. Thuan, ``Facial expression recognition using
  deep convolutional neural networks,'' \emph{International Conference on
  Knowledge and Systems Engineering}, 2017.

\bibitem{Xu:2015:FERBasedOn}
M.~Xu, W.~Cheng, Q.~Zhao, L.~Ma, and F.~Xu, ``Facial expression recognition
  based on transfer learning from deep convolutional networks,'' in \emph{11th
  International Conference on Natural Computation}, Aug 2015, pp. 702--708.

\bibitem{Jung:2015:DevelopmentOf}
H.~Jung, S.~Lee, S.~Park, B.~Kim, J.~Kim, I.~Lee, and C.~Ahn, ``Development of
  deep learning-based facial expression recognition system,'' in \emph{21st
  Korea-Japan Joint Workshop on Frontiers of Computer Vision}, Jan 2015, pp.
  1--4.

\bibitem{Zavarez:2017:CrossDatabase}
M.~V. Zavarez, R.~F. Berriel, and T.~Oliveira-Santos, ``Cross-database facial
  expression recognition based on fine-tuned deep convolutional network,'' in
  \emph{30th SIBGRAPI Conference on Graphics, Patterns and Images}, Oct 2017,
  pp. 405--412.

\bibitem{Valero:2016:AutomaticFER}
H.~G. Valero, ``Automatic facial expression recognition,'' 2016.

\bibitem{Zhang:2016:ADeepNN}
T.~Zhang, W.~Zheng, Z.~Cui, Y.~Zong, J.~Yan, and K.~Yan, ``A deep neural
  network-driven feature learning method for multi-view facial expression
  recognition,'' \emph{IEEE Transactions on Multimedia}, vol.~18, no.~12, pp.
  2528--2536, Dec 2016.

\bibitem{Alizadeh:2017:CNNForFER}
S.~Alizadeh and A.~Fazel, ``Convolutional neural networks for facial expression
  recognition,'' \emph{arXiv preprint arXiv:1704.06756}, 2017.

\bibitem{Barsoum:2016:TrainingDeep}
E.~Barsoum, C.~Zhang, C.~C. Ferrer, and Z.~Zhang, ``Training deep networks for
  facial expression recognition with crowd-sourced label distribution,'' in
  \emph{Proceedings of the 18th ACM International Conference on Multimodal
  Interaction}, ser. International Conference on Multimodal Interaction
  2016.\hskip 1em plus 0.5em minus 0.4em\relax New York, NY, USA: ACM, 2016,
  pp. 279--283.

\bibitem{Raghuvanshi:2016:FER}
A.~Raghuvanshi and V.~Choksi, ``Facial expression recognition with
  convolutional neural networks,'' 2016.

\bibitem{Dufourq:2017:EDEN}
E.~Dufourq and B.~A. Bassett, ``Eden: Evolutionary deep networks for efficient
  machine learning,'' in \emph{2017 Pattern Recognition Association of South
  Africa and Robotics and Mechatronics}, Nov 2017, pp. 110--115.

\bibitem{Zhou:2016:FERBased}
S.~Zhou, Y.~Liang, J.~Wan, and S.~Z. Li, \emph{Facial Expression Recognition
  Based on Multi-scale CNNs}.\hskip 1em plus 0.5em minus 0.4em\relax Cham:
  Springer International Publishing, 2016, pp. 503--510.

\bibitem{AlShabi:2016:FER}
M.~Al{-}Shabi, W.~P. Cheah, and T.~Connie, ``Facial expression recognition
  using a hybrid cnn-sift aggregator,'' \emph{CoRR}, vol. abs/1608.02833, 2016.

\bibitem{Rassadin:2017:GroupLevelER}
A.~Rassadin, A.~Gruzdev, and A.~Savchenko, ``Group-level emotion recognition
  using transfer learning from face identification,'' in \emph{Proceedings of
  the 19th ACM International Conference on Multimodal Interaction}, ser.
  International Conference on Multimodal Interaction 2017.\hskip 1em plus 0.5em
  minus 0.4em\relax New York, NY, USA: ACM, 2017, pp. 544--548.

\bibitem{Lopes:2017:FER}
A.~T. Lopes, E.~de~Aguiar, A.~F.~D. Souza, and T.~Oliveira-Santos, ``Facial
  expression recognition with convolutional neural networks: Coping with few
  data and the training sample order,'' \emph{Pattern Recognition}, vol.~61,
  pp. 610 -- 628, 2017.

\end{thebibliography}

\end{document}